\documentclass{article}

    \PassOptionsToPackage{numbers, compress}{natbib}

    \usepackage[final]{neurips_2020_ml4ps}

\usepackage[utf8]{inputenc} 
\usepackage[T1]{fontenc}    
\usepackage{hyperref}       
\usepackage{url}            
\usepackage{booktabs}       
\usepackage{amsfonts}       
\usepackage{nicefrac}       
\usepackage{microtype}      

\usepackage{subcaption}
\usepackage{multirow}
\usepackage{lineno}
\usepackage{xcolor}
\usepackage{graphicx}
\usepackage{amsmath}
\usepackage{amssymb}

\usepackage[symbol]{footmisc}
\definecolor{RoyalBlue}{RGB}{65, 105, 225}

\title{Optical Wavelength Guided Self-Supervised Feature Learning For Galaxy Cluster Richness Estimate}

\author{Gongbo~Liang\textsuperscript{1,2}\hspace{.75em}
		Yuanyuan~Su\textsuperscript{1}\hspace{.75em}
        Sheng-Chieh~Lin\textsuperscript{1}\hspace{.75em}
		Yu~Zhang\textsuperscript{1}\\
		\textbf{
		Yuanyuan~Zhang\textsuperscript{3}\hspace{.75em}
		Nathan~Jacobs\textsuperscript{1}} \\ [1ex]
		$^1$ University of Kentucky, Lexington, KY, USA~~\\
		$^2$ Eastern Kentucky University, Richmond, KY, USA~~\\
		$^3$ Fermi National Accelerator Laboratory, Batavia, IL, USA \\ [1ex]
		Project Page: \href{http://www.gb-liang.com/OWG}{www.gb-liang.com/owg}
		}

\begin{document}

\maketitle

\begin{abstract}
    Most galaxies in the nearby Universe are gravitationally bound to a cluster or group of galaxies. Their optical contents, such as optical richness, are crucial for understanding the co-evolution of galaxies and large-scale structures in modern astronomy and cosmology. The determination of optical richness can be challenging. We propose a self-supervised approach for estimating optical richness from multi-band optical images. 
    The method uses the data properties of the multi-band optical images for pre-training, which enables learning feature representations from a large but unlabeled dataset. 
    We apply the proposed method to the Sloan Digital Sky Survey. The result shows our estimate of optical richness lowers the mean absolute error and intrinsic scatter by $11.84\%$ and $20.78\%$, respectively, while reducing the need for labeled training data by up to $60\%$. 
    We believe the proposed method will benefit astronomy and cosmology, where a large number of unlabeled multi-band images are available, but acquiring image labels is costly. 
\end{abstract}

\section{Introduction}

Most of the galaxies in the Universe, including our own galaxy, reside in clusters or groups of galaxies~\cite{springel2003history}. The optical richness of a galaxy cluster, $\lambda$, is a measure of the number of galaxies physically bounded to the system. It can be a tracer of the underlying dark matter halo that can be utilized to constrain dark energy parameters~\cite{haiman2001constraints}. 
$\lambda$ 
is also crucial to our understanding of the galaxy evolution and the growth of large structures in the Universe~\cite{luparello2015brightest}. However, the measurement of $\lambda$ is confronted by the presence of foreground and background objects, leading to the uncertainty of the location of a given galaxy along the line of sight. The membership of cluster galaxies can be determined with their spectroscopic redshifts. But spectroscopic surveys are expensive, and it is yet to be practical to perform spectroscopic follow-up for a large fraction of the sky~\cite{levi2019dark}. 

\begin{figure*}[!h]
\centering
\renewcommand\thefigure{S1}
    \begin{subfigure}[b]{0.175\textwidth}
        \centering
        \includegraphics[width=\textwidth]{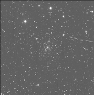}
        \caption{\textit{g}-band}
        \label{fig:dataset_1}
    \end{subfigure}~~
    \begin{subfigure}[b]{0.175\textwidth}
        \centering
        \includegraphics[width=\textwidth]{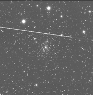}
        \caption{\textit{i}-band}
        \label{fig:dataset_1}
    \end{subfigure}~~
    \begin{subfigure}[b]{0.175\textwidth}
        \centering
        \includegraphics[width=\textwidth]{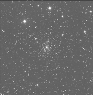}
        \caption{\textit{r}-band}
        \label{fig:dataset_1}
    \end{subfigure}~~
    \begin{subfigure}[b]{0.175\textwidth}
        \centering
        \includegraphics[width=\textwidth]{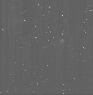}
        \caption{\textit{u}-band}
        \label{fig:dataset_1}
    \end{subfigure}~~
    \begin{subfigure}[b]{0.175\textwidth}
        \centering
        \includegraphics[width=\textwidth]{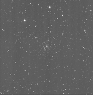}
        \caption{\textit{z}-band}
        \label{fig:dataset_1}
    \end{subfigure}
    \caption{A galaxy cluster shows in multi-band images.}
\label{fig:multi-band}
\end{figure*}

Photometric surveys in multiple wavelength ranges, i.e. multi-band optical imaging, are much more achievable and have become 
an advanced imaging technique that is widely used for astronomical and cosmological studies~\cite{nyland2017application}. Each multi-band optical image is a single channel image that is acquired using a specific optical wavelength band. The commonly used wavelength bands include \textit{u}, \textit{g}, \textit{r}, \textit{i}, and \textit{z} (see Figure~\ref{fig:multi-band} for an example). The imaging modality is efficient for detecting galaxy clusters, but the richness of a galaxy cluster cannot be directly estimated. 

Convolutional neural networks (CNNs) have been rapidly adopted in the astronomy and cosmology domains~\cite{ntampaka2019deep,su2020machine,zhang2021multi}. However, training a CNN typically requires a large number of labeled examples~\cite{mihail2019automatic,zhang2019defense,liang2020imporved,salem2020learning}, limiting their applicability for many real-world problems~\cite{litjens2017survey,yu2019clinical,wang2020inconsistent}. To overcome this limitation, we propose a novel self-supervised training method for $\lambda$ estimation. The proposed method utilizes the data properties of the multi-band optical images and enables model training with a large but unlabeled dataset. 

We believe rich information about $\lambda$ is embedded in different optical bands. Thus, we exploit latent connections between optical bands and $\lambda$ by learning the feature representations through optical wavelength band classifications. More specifically, we first learn image feature through band classification tasks. No manual annotations are required for the feature learning since wavelength band labels are known. Next, a downstream $\lambda$ estimation network is built using the pre-trained feature extractor and trained on a small labeled dataset. 
We apply the proposed method to images taken from the Sloan Digital Sky Survey (SDSS)~\cite{blanton2017sloan}. The result shows that our method significantly improves the $\lambda$ estimate while reducing the need for labeled training data by up to $60\%$.

\section{Method}
The proposed method contains two branches (Figure~\ref{fig:architecture}): 1) an optical wavelength-guided feature learning branch and 2) a galaxy cluster richness estimation branch. A CNN feature extractor is shared between the two branches. The two branches can be optimized simultaneously or trained in separate phases. For simplicity, we present the two branches in a two-phase training setup.

\subsection{Feature Learning}
\label{sec:feature_learning}
\begin{figure*}[!tb]
    \centering
    \includegraphics[width=0.825\textwidth]{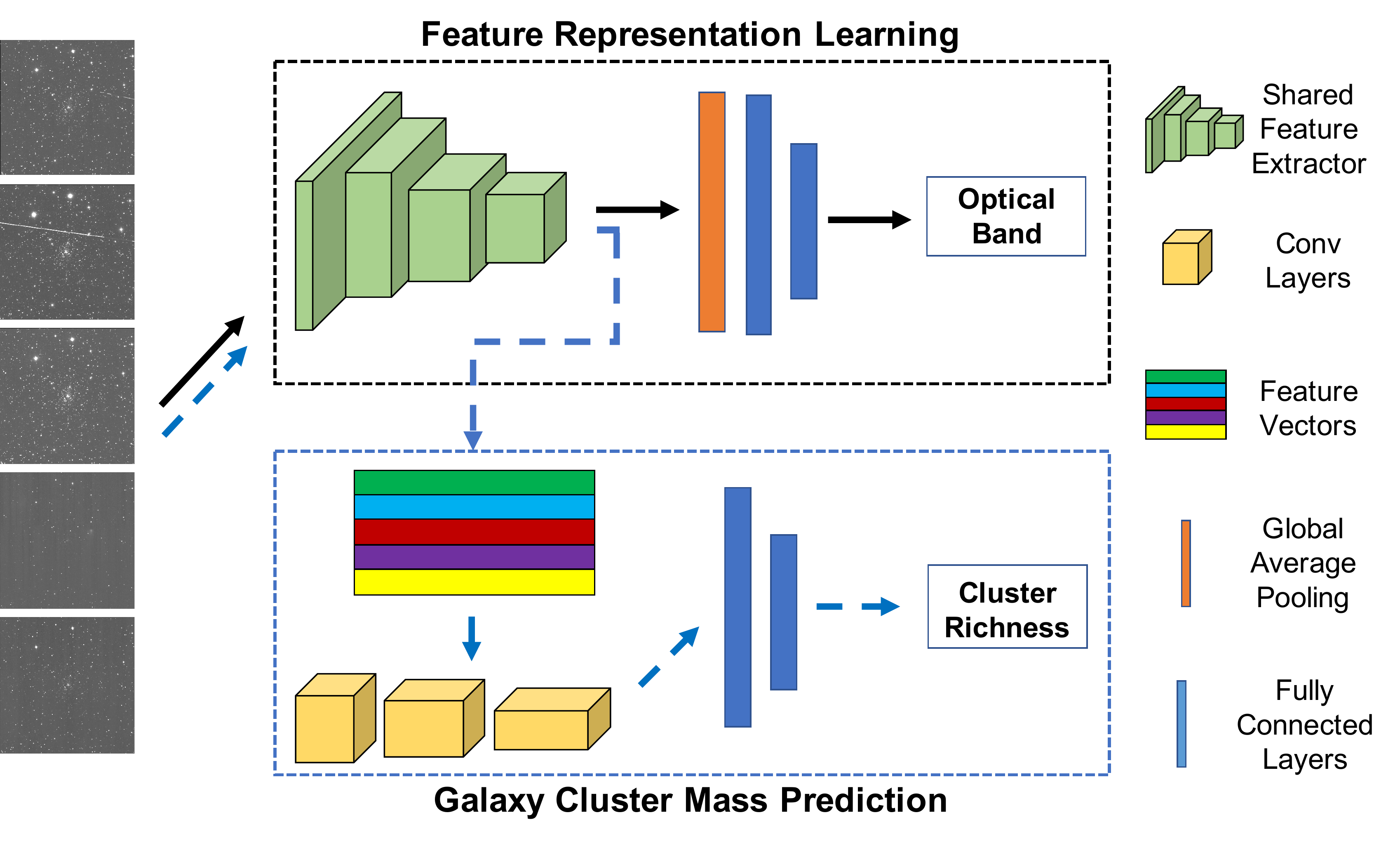}
    \caption{Two branches self-supervised learning architecture: 1) feature extractor via optical band classifications (solid black line); 2) galaxy cluster richness estimation (dashed blue line).}
\label{fig:architecture}
\end{figure*}

The feature learning branch is a classification network with multiple convolutional (Conv) layers followed by a global average pooling (GAP) layer and two fully connected (FC) layers (Figure~\ref{fig:architecture} solid black line). The network treats each of the multi-band images independently and predicts the corresponding optical band label for each image. 
The Conv layers are used as a feature extractor, which learns meaningful features from the images. The FC layers output the logits for a softmax. Cross-entropy loss is used for the feature learning branch training.
Since the band label is automatically assigned to each image when the image was acquired, no manually annotated label is needed when training the feature learning branch. Besides, by treating each optical image independently, the feature learning branch naturally increases the training set size by a factor of five. 

\subsection{Galaxy Cluster Richness Estimation}
The richness estimation branch is another CNN network (Figure~\ref{fig:architecture} dashed blue line), which shares the feature extractor with the feature learning branch. The network takes multi-band optical images as input and estimates the galaxy cluster richness ($\lambda$). Five multi-band optical images of the same galaxy cluster are passed through the feature extractor. The five feature maps are then concatenated and passed through a sequence of Conv layers and FC layers.  
The Conv layers aim to 1) convert the band classification trained features to the regression task, and 2) learn meaningful representations across the five different bands. The FC layers are used to perform the regression prediction. Table~\ref{table:code_mass_estimate} shows the pseudocode of training the richness estimation branch. 

The feature extractor can be used fixed or jointly optimized during the richness estimation branch training. The mean squared error (MSE) loss with the intrinsic scatter between the ground-truth richness and predicted richness (a custom regularization term) is used in the training. The loss can be written as:
\begin{equation}
\text{loss}=\frac{1}{n}\sum_{i=1}^{n}(y_i - \hat{y_i})^2 + \alpha \text{std}(|Y-\hat{Y}|), 
\label{eq:loss}
\end{equation}
where the first term is the MSE loss and the second term is the intrinsic scatter between the ground-truth richness and predicted richness, in which $\alpha$ is a weight scalar, $\text{std}(\cdot)$ denotes the standard deviation, $|\cdot|$ denotes the absolute value, $Y$ and $\hat{Y}$ are the sets of ground-truth and the predicted $\lambda$ with the magnitude of $n$, $y_i$ and $\hat {y_i}$ indicate the ground-truth and predicted $\lambda$ for the $i^{th}$ sample in one training batch.

\subsection{Implementation}
\label{sec:implementation}
We implement this work in PyTorch~\cite{paszke2019pytorch}. The ResNet-18~\cite{he2016deep} model is used as the backbone of the feature learning branch. A $1\times 1$ Conv layer followed with BatchNorm~\cite{ioffe2015batch} and a rectified linear unit (ReLU) is added before the GAP layer of the ResNet-18 model. An FC layer with 512 neurons is added after the GAP layer. The SGD optimizer with a learning rate of~$10^{-4}$ and a momentum of $0.9$ is used in training. Cross-entropy is used as the loss function in feature learning branch training. 

The richness estimation branch contains the shared feature extractor, three Conv layers, and three FC layers. GAP is applied to each of the feature maps with an output shape of~$512\times1$. The five multi-band optical images of a galaxy cluster are passed through the feature extractor. Then, the five feature maps are concatenated and passed through the rest of the network. Table~\ref{table:branch_2} shows the detailed architecture of the richness estimation branch. The SGD optimizer with a learning rate of~$5\times10^{-6}$ and a momentum of $0.9$ is used in training. Equation~\ref{eq:loss} is used as the loss function for the richness estimation branch training. 

\section{Experiments}
\subsection{Experimental Setup}
We use the SDSS Release 12 dataset~\cite{alam2015eleventh} in this study. 
SDSS is an optical wide-field imaging survey covering one-third of the sky ($\sim14000\text{deg}^2$) and provides five broad-band data ($u$, $g$, $r$, $i$, $z$) with a typical depth of $\sim21$ magnitude in the $i$-band.
We select the images according to the cluster catalog generated by the red-sequence Matched-filter Probabilistic Percolation (redMaPPer) cluster finding algorithm~\cite{rykoff2014redmapper}. We set the image size to be the physical size of  $1$~Mpc~($=3.09\times10^{22}$m) around each cluster center.
The labeled values of $\lambda$ are taken from the results given by the redMaPPer algorithm. 

In total,~$24,596$ optical observations are used in this study.
In the feature learning stage, we randomly partition the dataset into training and testing sets, with a $4:1$ ratio.
We pre-train the feature learning branch for $100$ epochs. 
In the galaxy cluster richness estimation training stage, we randomly partition the dataset into ten sets of roughly equal size.
Ten-fold cross-validation is used to evaluate model performance. 
For both partitioning steps, we ensure that all images of a given galaxy cluster are in the same partition

\subsection{Evaluation Method}
We evaluate the proposed method based on the $\lambda$ estimation and the degree of need for labeled instances for training. We compare the proposed method (denoted as \textit{Ours}) against a baseline model (denoted as \textit{Base}). Both \textit{Ours} and \textit{Base} have the same architecture, but the feature extractor of \textit{Ours} is pre-trained via the optical band classification task. 
We train each model multiple times using different percentages of the training data (between $1\%$ and $100\%$). The data are randomly selected from each data fold. For the models that are trained with the same amount of training data, we ensure to use the same subsets in the training of both \textit{Ours} and \textit{Base}.
We use the mean absolute error (MAE) and the intrinsic scatter between the ground-truth richness and predicted richness (Sigma) as the evaluation metrics. For both metrics, a smaller value indicates a better performance. 

\subsection{Richness Estimate}
\begin{figure}[!tb]
\centering
\renewcommand\thefigure{S2}
    \begin{subfigure}[b]{0.435\textwidth}
        \centering
        \includegraphics[width=\textwidth]{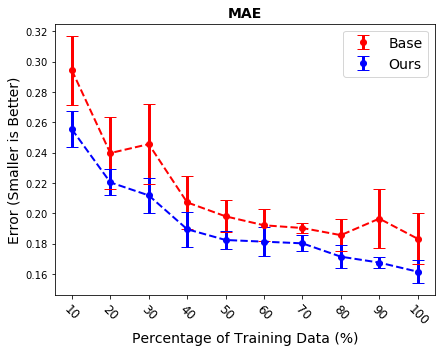}
    \end{subfigure}~~~~~~~~~~~~
    \begin{subfigure}[b]{0.435\textwidth}
        \centering
        \includegraphics[width=\textwidth]{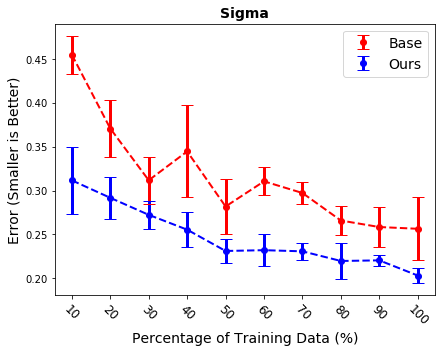}
    \end{subfigure}
    \caption{The performance (mean and standard deviation) of Base and Ours using $10\%$ to $100\%$ of training data. Left: The mean absolute error (MAE). Right: The intrinsic scatter between the ground-truth richness and predicted richness(Sigma).}
\label{fig:result}
\end{figure}

Table~\ref{table:result} and Figure~\ref{fig:result} show the results of \textit{Ours} and \textit{Base} using different amounts of training data. Each model was run three times. The mean value of the three trials are shown in Table~\ref{table:result}. The mean value with the standard deviation are shown in Figure~\ref{fig:result}. The result reveals that our approach is superior to the \textit{Base} model in all setting, with better gains when only a few labeled images are available. For instance, when using $1\%$ of the labeled data ($\approx$~245 instances), the \textit{Base} model has an MAE of $2.3923$ and a Sigma of $1.3304$, while the \textit{Ours} has $0.4188$ and $0.5166$, respective. The proposed method reduces the MAE by $82.49\%$ and reduces the Sigma by $61.70\%$.

\begin{table*}[!tb]
	\centering
	\addtolength{\tabcolsep}{-3.5pt}
	\small
	\caption{Detailed Performance of \textit{Base} and \textit{Ours}}
    \renewcommand{\arraystretch}{1.25}
    \addtolength{\tabcolsep}{-1.pt}
	\begin{tabular}{|c|c|c|c|c|c|c|c|c|c|c|c|c|c|}
    \hline
    \multirow{2}{*}{\textbf{Metric}} & \multirow{2}{*}{\textbf{Model}} & \multicolumn{12}{c|}{\textbf{Percentage of Training Data}}  \\
 \cline{3-14}
    &  & {$\bf{1\%}$}  & {$\bf{5\%}$} & {$\bf{10\%}$} & {$\bf{20\%}$} & {$\bf{30\%}$} & {$\bf{40\%}$} & {$\bf{50\%}$} & {$\bf{60\%}$} & {$\bf{70\%}$} & {$\bf{80\%}$} & {$\bf{90\%}$} & {$\bf{100\%}$} \\ \hline

    \multirow{2}{*}{MAE} 
    & Base 
    & $2.3923$ & $0.4792$ & $0.2943$ & $0.2397$ & $0.2456$ & $0.2072$ & $0.1980$ & $0.1921$ & $0.1903$ & $0.1856$ & $0.1965$ & $0.1832$ \\ \cline{2-14}
    & Ours
    & $\textbf{0.4188}$ & $\textbf{0.2566}$ & $\textbf{0.2554}$ & $\textbf{0.2204}$ & $\textbf{0.2118}$ & $\textbf{0.1895}$ & $\textbf{0.1824}$ & $\textbf{0.1813}$ & $\textbf{0.1802}$ & $\textbf{0.1714}$ & $\textbf{0.1675}$ & $\textbf{0.1615}$ \\ \hline
      
    \multirow{2}{*}{Sigma} 
    & Base 
    & $1.3304$ & $0.5810$ & $0.4549$ & $0.3708$ & $0.3118$ & $0.3447$ & $0.2819$ & $0.3106$ & $0.2973$ & $0.2659$ & $0.2585$ & $0.2565$ \\ \cline{2-14}
    & Ours
    & $\textbf{0.5166}$ & $\textbf{0.3318}$ & $\textbf{0.3117}$ & $\textbf{0.2918}$ & $\textbf{0.2724}$ & $\textbf{0.2555}$ & $\textbf{0.2312}$ & $\textbf{0.2321}$ & $\textbf{0.2308}$ & $\textbf{0.2198}$ & $\textbf{0.2205}$ & $\textbf{0.2032}$ \\ \hline
  \end{tabular}
  \label{table:result}
\end{table*}

The best performance of \textit{Base} is $0.1832$ MAE and $0.2565$ Sigma, when it is trained with $100\%$ of the labeled data. \textit{Ours} surpasses the best performance on MAE of \textit{Base} with only $50\%$ of training data ($0.1824$ MAE) and the best performance on Sigma of \textit{Base} with only $40\%$ of training data ($0.2555$ Sigma). Thus, the need for manual labels is reduced by $50\%$ or $60\%$ when using the proposed method. 
One particular observation is that for every fraction of the full training dataset considered, the proposed method is superior across all measures. The best performance of \textit{Ours} is $0.1659$ MAE and $0.2080$ Sigma, which are $11.84\%$ and $20.78\%$ improvement compared with \textit{Base}.

\subsection{Pre-Trained Feature}
CNN feature extractors encode images into a high dimensional space. For instance, the ResNet-18 feature extractor used in this work encodes an input image of $224\times 224\times 3$ pixels to a $7\times7\times512$ feature space~\cite{he2016deep}. Evaluation or analysis of the learned features is usually non-trivial due to the high dimensionality. We evaluate the feature extractor by applying an occlusion test on the pre-trained optical band classification network (solid black line in Figure~\ref{fig:architecture}). 

\begin{figure*}[!tb]
    \centering
    \includegraphics[width=0.985\textwidth]{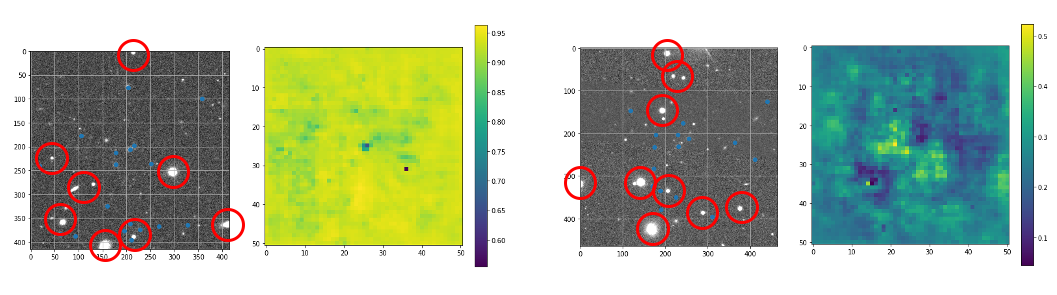}
    \caption{Two randomly selected occlusion testing results show that the optical band classification network may consider the member galaxies and foreground stars when making the decision. For each example, left: an optical image, right: occlusion map, blue dot: member galaxy, red circle: foreground star.}
    \label{fig:occlusion}
\end{figure*}

Figure~\ref{fig:occlusion} shows the occlusion testing results for two randomly selected samples. For each example, the input optical image is displayed on the left, and the occlusion map is displayed on the right. Blue dots indicate the member galaxies of the cluster taken from~\cite{rykoff2014redmapper}. Red circles indicate foreground stars.
To conduct this experiment, we use a small patch to occlude part of the input image and use the occluded image to test the model. We repeat this process by occluding every possible location of the input image. The pixel value of the occlusion map is the probability of being the correct prediction when the corresponding part is occluded. The brighter color indicates higher probability and darker color indicates lower probability. For instance, in Figure~\ref{fig:occlusion} Left, the occlusion map shows that when occluding the center of the image, the predicted probability changed dramatically, indicating cluster central regions may be more important than other areas to the decision-making process. 

By comparing the occlusion map and the input image, we noticed that a significant performance change often happens when occluding foreground stars or member galaxies. This systematic phenomenon may indicate the foreground stars and member galaxies are critical to the classification decision. From this end, we believe the pre-trained features may represent both foreground stars and member galaxies well. 

Conceptually speaking, galaxy richness estimation is a process of separating member galaxies from other objects in the same image. The ability to represent member galaxies and foreground stars is an important criterion that leads to the success of this task. Thus, we believe the feature learned by the proposed method is highly relevant to richness estimations.




\section{Concluding Remarks}
We proposed a novel self-supervised learning method for galaxy cluster richness estimation using multi-band optical images. The method utilizes the data properties of the multi-band optical images for pre-training and enables learning feature representations using a large but unlabeled dataset. We believe the proposed self-supervised feature learning method is not limited to galaxy cluster richness estimation. It is potentially useful for any task where multi-band images are available but acquiring manual labels is expensive and time-consuming.


\section*{Broader Impact}
A robust deep neural network usually requires a large labeled data set for training, which often does not exist in the real world. The astronomy and cosmology domains are not exceptions. Limited labeled data prevents the adoption of the latest advanced neural network techniques in the domains. We consider our contributions to this work as the following:

\begin{itemize}
\item A novel self-supervised approach on galaxy cluster richness estimation, which improves the galaxy cluster richness estimate while reducing the need for labeled training data.
\item The concept of using color bands as a guidance for pre-training is not limited to the astronomy and cosmology domains. It also works in the natural imaging domain (see Section~\ref{sec:natural_imaging}).
\item To our best knowledge, this is the first work that applies a self-supervised training strategy in galaxy cluster richness estimation.
\end{itemize}

\begin{ack}
This work was sponsored by Grant No. IIS-1553116 from the U.S. National Science Foundation. 

Funding for SDSS-III has been provided by the Alfred P. Sloan Foundation, the Participating Institutions, the National Science Foundation, and the U.S. Department of Energy Office of Science. The SDSS-III web site is http://www.sdss3.org/.
SDSS-III is managed by the Astrophysical Research Consortium for the Participating Institutions of the SDSS-III Collaboration including the University of Arizona, the Brazilian Participation Group, Brookhaven National Laboratory, Carnegie Mellon University, University of Florida, the French Participation Group, the German Participation Group, Harvard University, the Instituto de Astrofisica de Canarias, the Michigan State/Notre Dame/JINA Participation Group, Johns Hopkins University, Lawrence Berkeley National Laboratory, Max Planck Institute for Astrophysics, Max Planck Institute for Extraterrestrial Physics, New Mexico State University, New York University, Ohio State University, Pennsylvania State University, University of Portsmouth, Princeton University, the Spanish Participation Group, University of Tokyo, University of Utah, Vanderbilt University, University of Virginia, University of Washington, and Yale University.
\end{ack}

{\small
\bibliographystyle{IEEEtran}
\bibliography{bibfile}
}
\newpage
\begin{flushleft}
\Large \textbf{Supplementary Materials}
\end{flushleft}

\begin{table}[!h]
    \centering
    \renewcommand\thetable{S2}
    \caption{Detailed Architecture for the Galaxy Cluster Richness Estimate Branch}
    \begin{tabular}{ccccc}
    \hline\noalign{\smallskip}
    \textbf{Layer}  && \textbf{Kernel Shape} && \textbf{Out Shape}\\
    \noalign{\smallskip}
    \hline
    \noalign{\smallskip}
    Shared Feature Extractor && -- && $512\times5\times1$
    \\\noalign{\smallskip}\hline\noalign{\smallskip}
    Conv1  && $1\times1$ && $512\times5\times1$\\
    BatchNorm && -- && --\\
    ReLU && -- && --  \\\noalign{\smallskip}\hline\noalign{\smallskip}
    Conv2  && $3\times3$ && $510\times5\times16$\\
    BatchNorm && -- && --\\
    ReLU && -- && --\\\noalign{\smallskip}\hline\noalign{\smallskip}
    Conv3  && $3\times3$ && $508\times5\times64$\\
    BatchNorm && -- && --\\
    ReLU && -- && --  \\\noalign{\smallskip}\hline\noalign{\smallskip}
    FC1  && -- && $1024$\\
    FC2  && -- && $512$\\
    FC3  && -- && $1$ \\
    \noalign{\smallskip}
    \hline
    \end{tabular}
    \label{table:branch_2}
\end{table}

\begin{table}[!h]
\begin{center}
\renewcommand\thetable{S1}
    \caption{Pseudocode for Galaxy Cluster Richness Estimation Branch Training}
    \begin{tabular}{l}
    \hline\noalign{\smallskip}
    \textbf{Pseudocode} \\
    \noalign{\smallskip} \hline \noalign{\smallskip}
    /\ /\  ~let \textit{CNN\textsubscript{ft}} be the shared feature extractork\\
    /\ /\  ~let \textit{CNN\textsubscript{reg}} be the regression network\\
    \noalign{\smallskip}\noalign{\smallskip}
    outputs $\leftarrow$ [] ~~ ~~ ~~/\ /\  ~an empty array \\
    for every galaxy cluster $C$\{ \\
    ~~~~~~~~/\ /\ ~let $[I_h, I_i, I_r, I_u, I_z]$ be multi-band images of $C$ \\
    ~~~~~~~~ft $\leftarrow$ [] ~~ ~~ ~~/\ /\  ~an empty array \\
    ~~~~~~~~for(i=0; i$<$5, i++)\{ \\
    ~~~~~~~~~~~~~~~~ft[i].concatenate(\textit{CNN\textsubscript{ft}}($C$[i])) \\
    ~~~~~~~~\} \\
    ~~~~~~~~outputs.append(\textit{CNN\textsubscript{res}}(ft))\\
    \} \\
    \noalign{\smallskip}\noalign{\smallskip}\noalign{\smallskip}
    loss = loss\_function(outputs, ground\_truth) \\
    
    \noalign{\smallskip} \hline
    \end{tabular}
    \label{table:code_mass_estimate}
\end{center}
\end{table}

\renewcommand\thesection{S1}
\section{Performance on Natural Imaging Data}
\label{sec:natural_imaging}
We believe that the proposed method is not only limited to the astronomy and cosmology domains. 
We show the evaluation result of the proposed method in the natural domain using the CIFAR-10 and UCB200 datasets in this section. The self-training strategy in this section is slightly different from the one for galaxy cluster richness estimation, but they all follow the same principle of using color band/channel information for pre-training. 

\subsection{Pre-Trian Feature Extractor via Channel-Ordering}
In the natural imaging domain, images usually have three color channels: \textit{R}-channel, \textit{G}-channel, and \textit{B}-channel. For the pre-training task, instead of predicting the channel label of a single channel that is described in Section~\ref{sec:feature_learning}, we want to predict the color channel orders of a given input. More specifically, we randomly shuffle the color channels of an input image during the training time, such as from RGB to GBR. Then, we let the network predict the channel ordering of the given image. After the feature extractor is trained through the channel-ordering task, we can fine-tune it on the downstream classification tasks.

\subsection{Classification Result}
\begin{figure}[!h]
\centering
\renewcommand\thefigure{S3}
    \begin{subfigure}[b]{0.445\textwidth}
        \centering
        \includegraphics[width=\textwidth]{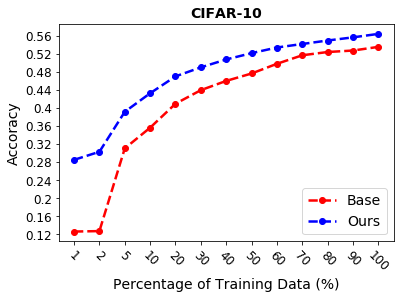}
        \label{fig:dataset_1}
    \end{subfigure}~~~~~~~~~~~~
    \begin{subfigure}[b]{0.445\textwidth}
        \centering
        \includegraphics[width=\textwidth]{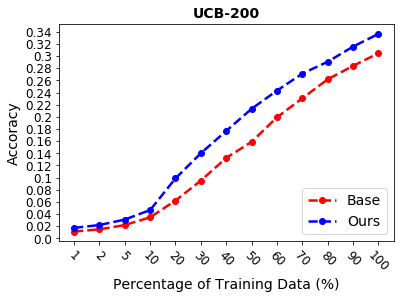}
        \label{fig:dataset_1}
    \end{subfigure}
    \caption{Channel-ordering pre-trained model performance on CIFAR10 and UCB200.}
\label{fig:channel_ordering}
\end{figure}

Figure~\ref{fig:channel_ordering} shows the comparing result of the proposed method (\textit{Ours}) and the baseline model (\textit{Base}) on the CIFAR-10 and UBC-200 datasets. Both \textit{Base} and \textit{Ours} have the same ResNet-18 architecture. The only difference is that the feature extractor of \textit{Ours} is pre-trained on the channel-ordering task. 
The figure reveals that the \textit{Ours} improves the classification performance on CIFAR-10 and UCB200 by $5.38\%$ and $10.43\%$, respectively. The \textit{Base} achieved its best performance on both datasets uses $100\%$ of the training data. The \textit{Ours} can achieve a similar performance using only $60\%$ or $80\%$ of the training data, respectively.

\end{document}